\documentclass[11pt,letterpaper]{article}
\usepackage[english]{babel}
\usepackage{amsmath}
\usepackage{amsfonts, mathrsfs}
\usepackage{amssymb}
\usepackage{verbatim}
\usepackage{graphicx}
\usepackage{color}
\usepackage{natbib}
\usepackage{url}
\usepackage{tikz}
\usepackage{pgflibraryarrows}
\usepackage{pgflibrarysnakes}
\usepackage{float}
\usepackage{graphicx}
\usepackage{multirow}
\usepackage{inputenc}
\usepackage{epsfig}
\usepackage{epstopdf}
\usepackage{verbatim}
\usepackage{enumerate}
\usepackage{subcaption}
\usepackage{algorithmic}
\usepackage{algorithm}
\allowdisplaybreaks

\numberwithin{equation}{section}
\newtheorem{theorem}{Theorem}[section]

\newtheorem{proposition}[theorem]{Proposition}
\newtheorem{lemma}[theorem]{Lemma}
\newtheorem{example}[theorem]{Example}

\newtheorem{definition}[theorem]{Definition}

\definecolor{mypink1}{rgb}{0.858, 0.188, 0.478}

\usepackage{setspace}
\allowdisplaybreaks

\DeclareMathOperator*{\argmin}{\rm argmin}

\usepackage{hyperref}
\title{Dynamic Learning with Frequent New Product Launches: \\A Sequential Multinomial Logit Bandit Problem}
\makeatletter
\renewcommand\@date{{%
  \vspace{-\baselineskip}%
  \large\centering
  \begin{tabular}{@{}c@{}}
    Junyu Cao\textsuperscript{1} \\
    \normalsize jycao@berkeley.edu
  \end{tabular}%
  \quad \quad
  \begin{tabular}{@{}c@{}}
    Wei Sun\textsuperscript{2} \\
    \normalsize sunw@us.ibm.com
  \end{tabular}

  \textsuperscript{1}Industrial Engineering and Operations Research, University of California, Berkeley\par
  \textsuperscript{2}IBM Research

}}
\makeatother
\begin{document}
\maketitle
\begin{abstract}
Motivated by the phenomenon that companies introduce new products to keep abreast with customers' rapidly changing tastes, we consider a novel online learning setting where a profit-maximizing seller needs to learn customers' preferences through offering recommendations, which may contain existing products and new products that are launched in the middle of a selling period. We propose a sequential multinomial logit (SMNL) model to characterize customers' behavior when product recommendations are presented in tiers. For  the offline version with known customers' preferences, we propose a polynomial-time algorithm and characterize the properties of the optimal tiered product recommendation. For the online problem, we propose a learning algorithm and quantify its regret bound. Moreover, we extend the setting to incorporate a constraint which ensures every new product is learned to a given accuracy. Our results demonstrate the tier structure can be used to mitigate the risks associated with learning new products.
~
\end{abstract}

\section{Introduction}
Facing increasingly savvy customers whose preferences are rapidly changing, 
companies that choose to play it safe by remaining with traditional product lines risk being overtaken by competitors more in tune with their customers. 
A coping strategy adopted by companies  is to frequently launch new products and learn from the  market responses. 
Between the tried-and-true existing products and new products with little or no history, companies face a dilemma - they have to offer new products in order to  understand the changing market dynamics so as to improve longer-term profitability, yet they may have to sacrifice short-term profitability. The central question is, how can a company quickly learn customers' preferences while mitigating the risks inherent in new products? 

We approach this question as an online learning task. We consider a seller whose goal is to maximize cumulative profit over a selling horizon $T$. She 
will introduce several new products at different times during the selling period.  For every customer, the seller determines some products to offer\footnote{We use ``recommend'' and ``offer'' interchangeably in this work. }, which may include  the existing and/or new products.   Based on the customer's response, the seller updates her belief on the latent customers' preferences (also known as product valuations), and uses the information to optimize the product selection for the next customer. As we will show in the paper, many new products with relatively low profit will never 
be offered from a pure profit-maximizing objective. 

In reality, companies often intentionally price new products low to gain exposure and to entice customers to give them a try. Thus, many new products may have relatively low profits, yet learning from these product is crucial for understanding customers' preferences, and enabling companies to make better business decisions in the future. To model such behavior, we impose a constraint, termed ``minimum learning criterion'', which requires \emph{every new} product to be offered and learned to a given accuracy. 
A direct implication is that the seller will be bearing additional cost of learning as she makes less money from these products. It is natural to ask what can be done to reduce such cost. 

We will show that a judicious choice of presenting products is capable of mitigating some costs associated with learning new products. In our setting, products are presented in tiers, indicating the precedence in which  customers discover them. 
For a given customer, a seller first offers the products on the first tier. If none are selected, the seller then presents the second tier, and so on.  Priorities are embedded in tiers as product placement affects product visibility. Such product offerings are ubiquitous in the online marketplace. For instance, when companies send multiple emails or app notifications to promote products, 
or on a website where products are displayed over multiple pages and customers have to take an action (such as clicking on ``Next'' or ''Load more'') to access the next set of products. 

To capture the customer's responses when recommendations come in tiers, we propose a sequential multinomial logit (SMNL) model, which generalizes the multinomial logit (MNL) model that has been extensively studied in the literature (e.g., \citealp{agrawal2017mnl, talluri2004revenue, train2009discrete}). Besides offering priorities in which products are being shown, we will prove additional benefits of tiered product recommendation, i.e., i) it is  capable of achieving higher profit than displaying all products at once, and ii) reducing the profit risks associated with new products.

In this paper, we refer to the online task of learning customers' preferences through tier-based product recommendations as the \emph{SMNL Bandit} problem. The contribution of our work is threefold:

\begin{enumerate}[1.]
    \item {We propose a novel SMNL model to capture customer's sequential choice behavior.} For the offline problem with known customers' preferences, we provide a polynomial-time algorithm to solve the profit-maximization  problem, and characterize the properties of the optimal tiered product offering. 
    \item For an online setting where new products are  frequently launched at different times during the selling horizon, we propose an online learning algorithm for the \emph{SMNL Bandit} problem, and characterize its regret bound.
    \item We extend the online setting to incorporate a constraint which  ensures all new products are learned to a given accuracy, and demonstrate how the tier structure in product presentation can be exploited to mitigate risks with new products.
      
\end{enumerate}

\section{Literature review}
The first stream of work that our paper is related to is assortment optimization. It refers to the problem of selecting a set of products to offer to a group of customers so as to maximize the revenue  when customers make purchases according to their preferences. It is a central topic in economics, marketing, and the operations management research literature. We refer the reader to \citealp{kok2008assortment} for a comprehensive review.
\citealp{talluri2004revenue} is the first paper that models customers' preferences with the MNL model for  the assortment planning problem. 
 \citealp{flores2018assortment} study the assortment optimization problem with a different sequential choice model known as the perception-adjusted Luce model and characterize the optimal assortment for the offline problem. {Besides the customers' preferences are modelled differently, }  we also study the problem in the online setting and investigate the learning policy with new products. 


Another related topic is the multi-armed bandit (MAB) problem (e.g., \citealp{robbins1985some}; \citealp{sutton1998reinforcement}). Our problem falls under the combinatorial setting (\citealp{chen2013combinatorial}) since the retailer's decision is a combination of different products. A naive approach is to treat each possible combination as an arm. However, the number of arms increases exponentially with the number of products with this approach. Other combinatorial bandit work assuming linear reward (\citealp{auer2002using, rusmevichientong2010linearly}) or independent rewards (\citealp{chen2013combinatorial}) cannot be directly applied to our model. Recent work on assortment optimization (such as \citealp{cheung2017thompson,  agrawal2017mnl, agrawal2017thompson, saure2013optimal, rusmevichientong2010dynamic}) extend the MNL assortment problem from the offline setting to  online  where customers' preferences are unknown a priori and need to be learned. 
Our work is more closely related to \citealp{agrawal2017mnl}, {but} with the following key differences. 
Firstly, we consider multi-tiered assortment. Despite their ubiquity in practice, there is little formal analysis in the literature on either the offline optimization problem or the online learning algorithms. Our work helps to bridge this gap. Secondly, we focus on learning in conjunction with new products launches, where we differentiate two cases depending on whether all new products need to be learned.   




\section{Problem formulation}

In this section, {we will formally set up our problem. We will first introduce the SMNL model, which describes the customers' behavior, and follow by formulating a profit maximization problem that the seller needs to solve.}

\subsection{Customer's behavior: SMNL model}

Discrete choice models such as the popular MNL model are derived under the assumption that a utility-maximizing customer chooses a product with the highest valuation among a available choice set $\bold{S}$ \cite{train2009discrete}. In a SMNL model, $\bold{S}$ consists of multiple tiers of products. 
For ease of notation, we will present a two-tier model where the choice set consists of two sets, i.e., $\bold{S}:=(S_1, S_2)$. We will refer to $S_1$ and $S_2$ as the priority tier and the secondary tier respectively, as products in $S_1$ enjoys greater visibility. Note that all our results can be generalized to incorporate more tiers. 

Customers arrive at discrete time $t = 1, \cdots, T$. For a customer arriving at time $t$, she is presented with a choice set $\bold{S}^t$ that is selected by the seller.   Under the SMNL model, a customer first considers products from the priority tier $S_1$. If none are selected, she will then consider the secondary tier $S_2$ and decide whether to select any product from $S_2$. Note that no-purchase is also one of the choices that the customer can make. The probability that a customer purchases product $i$ is denoted as $p_i(\bold{S})$ and no-purchase as $p_0(\bold{S})$, i.e.,
\begin{equation}
 p_i(\bold{S})=
\left\{
\begin{aligned}
&\frac{v_i}{1+\sum_{j\in S_1}v_j},& \text{ if } i\in S_1\\
&\frac{1}{1+\sum_{j\in S_1}v_j}\frac{v_i}{1+\sum_{j\in S_2}v_j},& \text{ if } i\in S_2\\
&\frac{1}{1+\sum_{j\in S_1}v_j}\frac{1}{1+\sum_{j\in S_2}v_j},&\text{ if } i=0\\
&0, &\text{ otherwise,}
\end{aligned}
\right.\notag
\end{equation}
where $v_i$ is the product valuation or customers' preference for product $i$, which is assumed to be less than 1. 
For a product from $S_1$, its purchase probability follows that of a standard MNL model. On the other hand, the probability of purchasing a product from  $S_2$,  is  {the joint probability of} two events, i.e., the customer has not selected {any }product from $S_1$ and the customer selects a product from $S_2$. 

\subsection{Seller's profit maximization problem}
Knowing  customers' purchase probability as $p_i(\bold{S})$ {when offering $\bold{S}$}, the seller needs to select a subset of products from all available products to form $S_1$ and $S_2$.  
We assume there are two pre-determined sets of product candidates, $X_1$ and $X_2$.  We want to point out that the two candidate sets need not be  mutually exclusive, and can completely overlap each other. A seller has the flexibility to assign products as candidates for the priority tier based on   sales, trendiness, {inventory}, and other business criteria.  

Denote the profit of product $i$ by $r_i$ and the profit obtained from $\bold{S}$ by $R(\bold{S})$. The expected profit 
can be expressed as 
$
E[R(\bold{S})]= \sum_{i\in S} r_i p_i(\bold{S})=\frac{\sum_{i\in S_1}r_iv_i}{1+\sum_{i\in S_1}v_i}+\frac{1}{1+\sum_{i\in S_1}v_i}\frac{\sum_{i\in S_2}r_iv_i}{1+\sum_{i\in S_2}v_i}.$
The seller's optimization problem is to  select two subsets of products $S_1$ and $S_2$ from the candidate sets $X_1$ and $X_2$ respectively. That is,  
\begin{align}
\max_{\bold{S}}\quad&  E[R(\bold{S})]\label{eq:optimization}\\
\text{s.t.} \quad & S_k\subseteq X_k, \quad \forall k\in\{1, 2\}.\nonumber
\end{align}
We use $\bold{S}^*=(S_1^*, S_2^*)$ to denote the optimal tiered product offering.   

\section{Characteristics of the optimal tiered product offering $\bold{S}^*$}\label{ss.optimal}
We begin this section with a simple example to compare a two-tiered product offering with its single-tiered counterpart. 

\textbf{Example 1. }
Suppose there are two products with profit $r_1=10, r_2=1$ and valuation $v_1=0.1, v_2=1$ respectively. The optimal one-tier recommendation is to offer both products simultaneously and the corresponding expected profit is given by
$E[R(\{1,2\})]=\frac{r_1v_1+r_2v_2}{1+v_1+v_2}=\frac{10*0.1+1*1}{1+0.1+1}=0.952.$
The optimal two-tier recommendation is to offer product 1 on the priority tier and product 2 on the secondary tier. The resulting profit 
$E[R((\{1\},\{2\}))]=\frac{10*0.1}{1+0.1}+\frac{1}{1.1}\frac{1*1}{1+1}=1.36 > E[R(\{1,2\})]$. 

This example shows that the tiered structure offers  flexibility in presenting products, which translates into higher profit. Intuitively, the tiered recommendation prioritizes  products with higher profits to be shown first. We can  formalize this observation by analyzing the seller's problem (\ref{eq:optimization}) in an offline setting where the product valuation $v_i$ is given.  

We now introduce two definitions which will help us characterize the properties of the optimal tiered product offering. 
\begin{definition}[Profit-ordered set] We call $S_k\subseteq X_k$ is a profit-ordered set if 
$\min_{i\in S_k} r_i\geq \max_{i\in X_k\backslash S_k} r_i,$ for $k\in \{1,2\}$. 
\end{definition}

\begin{definition}[Profit-ordered by tier] If there exist $i\in S_{1}$ and $j\notin S_{2}$ such that $r_i<r_j$, then $\bold{S}=(S_1, S_2)$ is not profit-ordered by tier. Otherwise, it is profit-ordered by tier. 
\end{definition}

\begin{example} 
Suppose $X_1=\{1,2,6,8\}$, $X_2=\{3,7,9,10\}$, with profit $r_i=i$ for all $i$, then the sets $\bold{S}=(\{6,8\},\{7,9,10\})$, $(\{8\},\{3,7,9,10\})$ are both profit-ordered by tier while the sets $\bold{S}=(\{2,6,8\},\{7,9,10\})$, $\{(6,8),(9,10)\}$ are not.
\end{example}

\begin{proposition}\label{L.candidate}
The optimal product offering $\bold{S}^*$ to the optimization problem (\ref{eq:optimization}) in each tier is a profit-ordered set. In addition, $\bold{S}^*$ is profit-ordered by tier.
\end{proposition}

Due to the space constraint, we only include proof sketches for the key results in the paper. All detailed proofs can be found in the supplementary material.

\emph{Proof sketch:} 
We show $S_1^*$ is profit-ordered by contradiction. Supposedly, there exists a $\bold{S}^*$ where $i\in S_1^*$ and $r_i< E[R(\bold{S}^*)]$, then we show that removing this product will increase the expected profit. Hence,  $\bold{S}^*$ is not optimal. A similar argument is used to show that if $i \notin S_1^*$, and $r_i> E[R(\bold{S}^*)]$, then adding it to the offering will increase the profit. Next, use the same argument to $S_2^*$ to obtain the desired result. 



To prove $\bold{S}^*$ is profit-ordered by tier, notice that the expected profit of $\bold{S}^*$ is at least as large as only offering $S_2$ since $S=(\emptyset, S_2)$ is also a feasible solution. Since we have shown that each tier in $\bold{S}^*$ is a profit-ordered set, i.e., for any $j\in S_1^*$, $r_j\geq E[R(\bold{S}^*)]$, and for any $i\notin S_2^*$, $r_i<E[R(S_2^*)]$. Therefore, $r_i\leq E[R(S_2^*)]\leq E[R(\bold{S}^*)]\leq r_j$ for any $i\notin S_2^*, j\in S_1^*$. This completes the proof.
$\blacksquare$

Proposition \ref{L.candidate} implies that a two-tier optimal recommendation can be characterized by a pair of profit thresholds $(\theta_1, \theta_2)$ with $\theta_1\geq \theta_2$, where $r_i\geq \theta_1$ and $r_j\geq \theta_2$ for any $i\in S_1$ and $j\in S_2$.  Therefore, the seller's optimization problem is polynomial-time solvable, as it follows directly from the fact that there are at most $|X_1||X_2|$  pairs of profit thresholds to enumerate through. In retail, as prices are discrete and often end with 9 or .99, there are far fewer unique price points than the number of products and the actual search space of profit thresholds is significantly smaller.

The profit-ordered structure of the optimal tiered recommendation provides important insights regarding the placement of a new product. We will generalize the result to a setting with multiple tiers. 
\begin{proposition}\label{C.multiple}
Denote the optimal recommendation before and after including a new product with profit $r_m$ to a candidate set as $\bold{S}^*=(S^*_1,S^*_2,\cdots, S^*_W)$ and $\hat{\bold{S}}^*$, respectively. Define $\bold{S}^*_j=(S^*_j,S^*_{j+1},\cdots, S^*_W)$. The following properties holds. 
\begin{enumerate}
   \item $E[R(\hat{\bold{S}}_j^*)]\geq E[R(\hat{\bold{S}}_{j+1}^*)]$ for any $j=1,\cdots, W-1$.
   \item If $E[R(\bold{S}^*_j)]< r_m< E[R(\bold{S}^*_{j-1})]$ for some $j$, then $m\in \hat{\bold{S}}^*$ but $m\notin \hat{S}_1^*\cup \hat{S}_2^*\cup \cdots\cup \hat{S}_{j-1}^*$. 
   \item If $r_m<E[R(S^*_W)]$, then $m\notin \hat{\bold{S}}^*.$
\end{enumerate} 
\end{proposition}
%
%
Proposition \ref{C.multiple} states that, for a two-tier product offering, unless a new product's profit is higher than $E[R(S_2^*)]$, where $S_2^*$ refers to what is currently being offered on the secondary tier, it will not be included. Therefore, this product will never be introduced or learned. As discussed in the introduction, many new products could have relatively low profit, but learning is crucial for providing insights to improve long-term profitability.  
This provides motivation for us to investigate an online learning task with a constraint to ensure all new products are learned to a given accuracy, which we will discuss in Section \ref{sect:newproducts}.

\section{Learning product valuations}\label{sect:online_learning}
In the previous section, we have assumed that valuations of products are known. In practice, these quantities are not given to the seller and have to be learned. 

\subsection{Online setup}\label{sect:onlinesetting}
We consider a general setting 
where $K$ new products are introduced at different time stamps during a selling horizon $T$. 
We allow several products to be launched at the same time. We use regret to measure the performance of a learning algorithm, where the regret for a policy  $\pi$  is defined as, 
$$Reg_\pi(T;\bold{v})=E_\pi\left[\sum_{t=1}^{T} R_t(\bold{S}^*,\bold{v})-R_t(\bold{S}^t,\bold{v})\right],$$
where $\bold{S}^*$ is the optimal tiered product offering when $\bold{v}$ is known, while $\bold{S}^t$ is the tiered recommendation offered to the customer arriving at time $t$. $R_t(\bold{S},\bold{v})$ denotes the profit accrued at time $t$ when offering recommendation $\bold{S}$.


For our learning task, we extend the framework in \citealp{agrawal2017mnl} which proposed a UCB-based algorithm for an online learning task with a MNL model. We want to {emphasize} that the tiered structure in the SMNL model significantly complicates the analysis  as the decisions across the tiers are \emph{interdependent}. Next, we will describe a counting process to derive an unbiased estimator of $v_i$ for $i\in \bold{S}$. 

\subsection{Unbiased estimator on product  valuation}
We divide the time horizon into epochs 
for the priority and the secondary tier respectively, i.e.,  $\mathcal{L}_1$ and $\mathcal{L}_2$. Let $\mathcal{L}=\mathcal{L}_1\cup \mathcal{L}_2$. In each epoch $l\in \mathcal{L}_k$ for $k=1,2$, we offer the same product selection $S_k^l$ for tier $k$ until a no-purchase in $S_k^l$ occurs. An epoch is labeled as $l$ if and only if $l$ epochs have been completed before $t$.
Let $\varepsilon_l^k$  contain all time steps during epoch $l$ when $S_k^l$ is shown to a customer. 

\begin{figure}
\centering
  \includegraphics[width=0.75\linewidth]{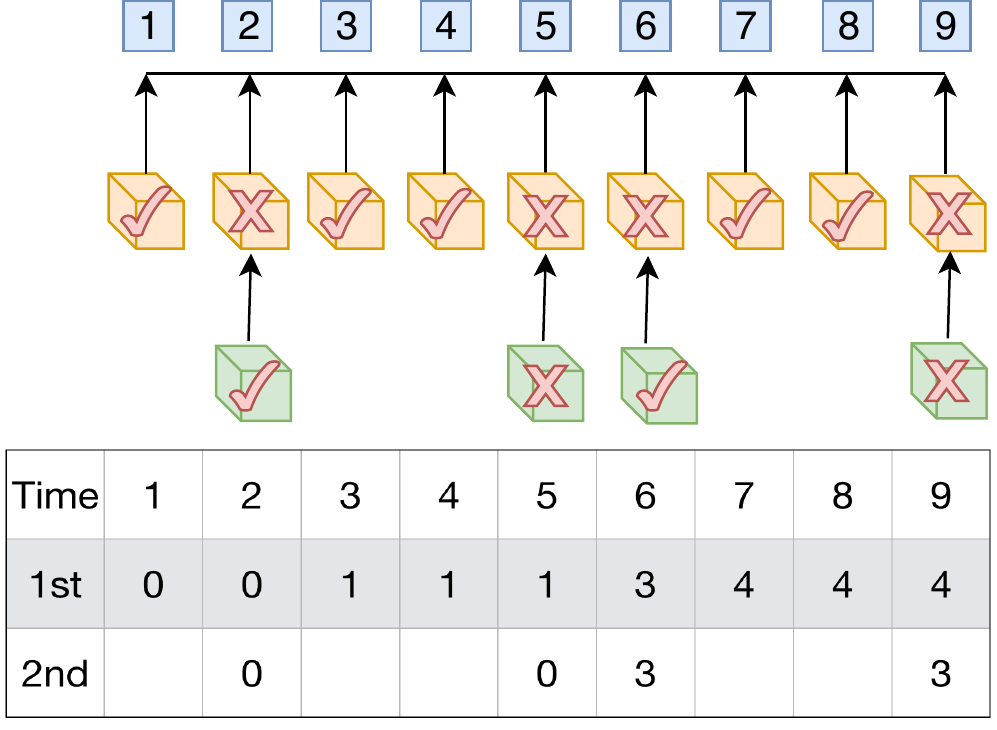}
  \caption{An illustrative example.}
  \label{fig:epoch}
\end{figure}

\begin{example}\
 Figure~\ref{fig:epoch} illustrates the counting process with an example, which shows the purchase decisions of 9 customers, i.e., $t=1,\cdots, 9$. The first customer selects a product from the priority tier, and the second customer selects a product from the secondary tier, and so on. The table in Figure~\ref{fig:epoch} shows how epochs are labeled for different tiers. Here we have $\mathcal{L}_1=\{0,1,3,4\}$ and $\mathcal{L}_2=\{0,3\}$. For $\mathcal{L}_1$, the epoch count at time $t$ is the same as the total number of no-purchases from both tiers before time $t$ . Thus, when $t=6$, epoch $l=3$ since there is a total of 3 no-purchases across both tiers by $t=5$. 
Note that for the secondary tier $k=2$, we only keep track of the time steps and the epoch count when $S^l_2$ is shown to a customer (i.e., the customer does not purchase any product from $S^l_1$). In terms of the time steps for each epoch, we have $\varepsilon_0^1=\{1,2\}$, $\varepsilon_1^1=\{3,4,5\}$, $\varepsilon_3^1=\{6\}$, $\varepsilon_4^1=\{7,8,9\}$, $\varepsilon_0^2=\{2,5\}$, and $\varepsilon_3^2=\{6,9\}$.
\end{example}

For any time step $t$, we use $c_t^k$ to denote the purchase decision of customer $t$ on tier $k$, i.e., $1(c_t^k=i)=1$  if the consumer purchased product $i\in S_k$, and 0 for a no-purchase. 
For any product $i\in S_1^l$ and $j\in S_2^l$, define 
$\hat{v}_{i,l}^{(1)}=\sum_{t\in\varepsilon_l^1}1(c_t^1=i)$ and $\hat{v}_{j,l}^{(2)}=\sum_{t\in\varepsilon_l^2}1(c_t^2=j)$
as the number of times a product $i$ is purchased in epoch $l$ as part of the primary or secondary tier selections respectively.

Let $\mathcal{T}_i^k(l)$  be the set of epochs which contain product $i$ in tier $k$ offering before epoch $l$. Define $T_i^k(l)=|\mathcal{T}_i^{k}(l)|$, which denotes the number of epochs which contain $i$ in tier $k$ offering before epoch $l$. Let $T_i(l)=T_i^1(l)+T_i^2(l)$, as the total number of epochs which contain $i$ in the tiered recommendation before epoch $l$. We compute $\bar{v}_{i,l}$ as the average number of times product $i$ is purchased per epoch, i.e.,
\begin{equation}\label{e.vbar}
\bar{v}_{i,l}=\frac{1}{T_i (l) }\left(\sum_{\tau\in\mathcal{T}^{1}_i(l)}\hat{v}_{i,\tau}^{(1)}+ \sum_{\tau\in\mathcal{T}^{2}_i(l)}\hat{v}_{i,\tau}^{(2)}\right).
\end{equation}


\begin{lemma}\label{unbiased_est}
$\hat{v}_{i,l}^{(k)}$ are i.i.d. geometric random variables with parameter $\frac{1}{1+v_i}$ for any $l$ and $k=1,2$. Therefore, they are unbiased i.i.d. estimators of $v_i$.
\end{lemma}

{\subsection{Learning algorithm for SMNL bandit}\label{sect:nocheapproduct}}
Define the upper confidence bound on $v_i$ as the follows, 
\begin{align}\label{e.vucb}
v_{i,l}^{UCB}:&=\bar{v}_{i,l}+\sqrt{\bar{v}_{i,l}\frac{48\log(K(l-l_{i,0})+1)}{T_i(l)}}\notag\\
&\hspace{5mm}+\frac{48\log(K(l-l_{i,0})+1)}{T_i(l)},
\end{align}
where $l_{i,0}$ is the initial launch epoch of product $i$, $\bar{v}_{i,l}$ is defined in Equation~\eqref{e.vbar}, and $K$ is the total number of products.

We briefly describe our UCB-based algorithm: In each epoch $l$, we use $\bold{v}_l^{UCB}$ to compute the optimal product offering . Denote $\tilde{\bold{S}}^l$ as the optimal product set when the value of products is $\bold{v}^{UCB}_l$ and $\bold{S}^*$ is the optimal set selected from the entire candidate sets including the new product. To bound the profit difference between $\bold{S}^*$ and $\tilde{\bold{S}}^l$, we derive the following result.

\begin{lemma}\label{L.upperbound}
Assume $0\leq v_i\leq v_i^{UCB}$ for all $i=1,\cdots, K$. Suppose $\bold{S}^*$ is an optimal tiered recommendation when the parameters of SMNL model are given by $\bold{v}$. Then $E[R(\bold{S}^*, \bold{v}^{UCB})]\geq E[R(\bold{S}^*,\bold{v})]$.
\end{lemma}

{Lemma \ref{L.upperbound} is a key step in the regret analysis for this UCB-based algorithm.} With Lemma \ref{L.upperbound}, on the ``large probability" event that $0\leq v_i\leq v_i^{UCB}$ for all $i=1,\cdots, K$, we can bound the difference $E[R(\bold{S}^*, \bold{v})]- E[R(\tilde{\bold{S}}^l,\bold{v})]$ by $E[R(\tilde{\bold{S}}^l, \bold{v}^{UCB})]- E[R(\tilde{\bold{S}}^l,\bold{v})]$. We will expand the regret analysis with more details in next section, where we impose an additional constraint to our learning task, as the current setting is a special case when the constraint is absent.



\section{Regret analysis with the minimum learning {criterion}}\label{sect:newproducts}

As we have discussed in Section \ref{ss.optimal}, 
by default a new product will only be included in the product offering if its profit $r_m\geq E[R(S_2)]$, where $S_2$ is the current product offering at the secondary tier. In other words, new products with profit $r_m< E[R(S_2)]$ will never be offered and and deprived of the learning opportunity. To have a more realistic setting, we will formally define a minimum learning constraint. We will then investigate a learning algorithm and quantify its resulting regret, starting with a single new product and later generalize to multiples.

\subsection{Minimum learning criterion}

We impose a constraint in our learning task to ensure that every product will be offered for at least a number of times to allow us to learn its valuation to a certain accuracy. More specifically, we require the estimated valuation $\bar{v}_{i}$ of every new product to be within $\epsilon$ to the true $v_i$ with a probability which is at least $1-\alpha$, where $\epsilon$ and $\alpha$ are two pre-determined parameters. We derive the following lemma which specifies the number of epochs $M$ needed to achieve a given level of estimation accuracy. 


\begin{lemma}[Minimum learning criterion]\label{L.Mvalue} For any $\epsilon$ and $\alpha>0$, if the number of epochs $M\geq \frac{192\log(2/\alpha+1)}{(-1+\sqrt{1+4\epsilon})^2}$, then $\bar{v}_i$ is within the $\epsilon$ confidence bound of $v_i$ with probability at least $1-\alpha$. That is,  $P(|\bar{v}_{i,l}-v_i|>\epsilon)<1-\alpha$
if $T_i(l)>\frac{192\log(2/\alpha+1)}{(-1+\sqrt{1+4\epsilon})^2}$.
\end{lemma}


We want to emphasize that
the constraint only affects a subset of new products which are otherwise excluded from being offered due to their relatively low profitability. Once they are offered and $M$ samples have been collected, they will be dropped out from future product recommendations. On the other hand, new products (along with some existing products) with {relatively} high profit will  continuously be offered after $M$ epochs and the estimation on their product valuations will be further improved. 
This is echoing what typically happens after product launches, where companies choose to continue or stop certain new products based on market response. 


\subsection{Learning with $r_m<E[R(S_2)]$}\label{ss.single}
In this section, we focus on with a setting when a single new product {with low profit} is launched in the middle of a selling horizon. Part of our goal is to determine the best way to include this product into learning. 


By Proposition \ref{C.multiple},  this low-profit product will be excluded from learning by default. In order to satisfy the minimum learning criterion, this new product will have to be offered for $M$ epochs, {where $M$ is determined by Lemma~\ref{L.Mvalue}}.  There are two possible strategies for us to learn this new product, i.e.,  either assigning it to {the priority tier or the secondary tier}.  

The answer to which is a better strategy is not immediately clear: While the duration of an epoch is shorter when a product is placed on the priority tier,  it could also mean that more of this product will be purchased. Hence, more profit loss and higher regret. On the other hand, even though a product placed on the secondary tier might make fewer sales, the duration of a single epoch could be much longer and the resulting regret could still be high since other products (in addition to the new product) also contribute to the total  regret. We now formally compare the two strategies by quantifying the corresponding regrets incurred during a single epoch. 

\textbf{Strategy 1: {Assigning new product to the priority tier}} \\Let $S_1'=S_1\cup \{m\}$, $S_2'=S_2$, and $\bold{S}'=(S_1', S_2')$. 
Let $N_1$ denote the number of times $S'_1$ has been shown to customers until a no-purchase occurs. {Note that $N_1$ follows the geometric distribution with mean $1+\sum_{j\in S_1'}v_j$, which depends on the valuation of all products in $S_1'$.}

Define the regret function during one epoch when the new product is included in the first tier as $G^{(1)}(\bold{S},\bold{v})$, i.e., 
\begin{align*}
G^{(1)}(\bold{S},\bold{v}):&=E\left[\sum_{t=1}^{N_1}R_t(\bold{S}^*,\bold{v})-R_t(\bold{S}',\bold{v})\right],
\end{align*}
where $\bold{S}=(S_1,S_2)$ and $\bold{S}'=(S_1', S_2)=(S_1\cup \{m\}, S_2)$. 


\textbf{Strategy 2: {Assigning new product to the secondary tier}}
\\Let $S_1''=S_1$, $S_2''=S_2\cup\{m\}$, and $\bold{S}''=(S_1'',S_2'')$. $N_2$ denotes the number of times $\bold{S}''$ has been shown to customers until a no-purchase from the entire product offering (i.e., both tiers). $N_2$ follows the geometric distribution with mean $(1+\sum_{j\in S_1} v_j)(1+\sum_{j\in S_2'}v_j)$.

Similarly, we define the corresponding regret function  as follows, 
\begin{align}\label{lem:opt_Q}
G^{(2)}(\bold{S},\bold{v}):=E\left[\sum_{t=1}^{N_2}R_t(\bold{S}^*,\bold{v})-R_t(\bold{S}'',\bold{v})\right],
\end{align}
where $\bold{S}=(S_1, S_2)$ and $\bold{S}''=(S_1, S_2')=(S_1, S_2\cup \{m\})$.  

 To compare the two strategies, we first need to determine the optimal action under a given strategy, then evaluate its ``best'' loss. The strategy which yields the lower regret is then considered a ``better'' strategy. 
Let $\bold{Q}^*$ and $\bold{Q}'^*$ denote the optimal solution that minimizes the regret $G^{(1)}$ and $G^{(2)}$, respectively, i.e.,  $\bold{Q}^*=\argmin_{\bold{S}} G^{(1)}(\bold{S},\bold{v})$ and $\bold{Q'}^{*}=\argmin_{\bold{S}} G^{(2)}(\bold{S},\bold{v})$. 
\begin{theorem}\label{L.better}
The optimal solution to $G^{(1)}$ and $G^{(2)}$ is the same as $\bold{S}^*$. That is, 
$\bold{Q}^*=\bold{Q}'^*=\bold{S}^*.$
In addition, 
we have
$G^{(1)}(\bold{S}^*,\bold{v})\geq G^{(2)}(\bold{S}^*,\bold{v}) = v_m(E[R(S_2^*)]-r_m).$
\end{theorem}

The implication of Theorem \ref{L.better}  is twofold. Firstly, it shows that the optimal offerings excluding the new product are identical for both strategies, irrespective of which tier the new product has been added to. In addition, they are also the same as the optimal offering $\bold{S}^*$ before the new product is added. In other words, there is no need to resolve the optimization problem with the added new product. Thus, it provides a simple learning algorithm for a new product with $r_m<E[R(S_2^*)]$: It is optimal to just add it to the secondary tier of the existing optimal product offering to satisfy the learning criterion.

Secondly, Theorem~\ref{L.better} also shows that with this optimal product offering $\bold{S}^*$, the regret is lower when the new product is added to the secondary tier. This result highlights the advantage of showcasing product recommendations in multiple tiers, in the sense we incur a smaller loss by displaying new products with higher risks (i.e., lower profit) on tiers with lower priorities.

\subsection{Learning with multiple new products}
This section focuses on a general setting similar to the one addressed in Section \ref{sect:onlinesetting}, except with the minimum learning constraint in place.  


We propose Algorithm~\ref{A.newproduct} to dynamically offer the recommendation which simultaneously explores and exploits. 
In Algorithm~\ref{A.newproduct}, for {each} epoch $l$, we compute the optimal tiered recommendation $\tilde{\bold{S}}^l$ given valuation $\bold{v}_l^{UCB}$. Based on Proposition \ref{L.better}, for any new product $i\in \mathcal{N}$ which is not included in $\tilde{\bold{S}}^l$, we add it to the second tier $\tilde{S}^l_2$. At the end of each epoch, we update $\bar{\bold{v}}_l$ and $\bold{v}^{UCB}_l$, which will be used to compute the recommendation for the next epoch.

\begin{algorithm}[t!]
   \caption{Exploration-Exploitation algorithm for SMNL-bandit with new products}\label{A.newproduct}
\begin{algorithmic}
   \STATE {\bfseries Initialization:} input $M, T$; $l=0$;
   \REPEAT
   \STATE  input product sets $X_1$ and $X_2$;
   \STATE  $\mathcal{N}=\{i:T_i^1(l)+T_i^2(l)<M\}$;
   \STATE   compute $\tilde{{\bf{S}}}^l$ given valuation ${\bf{v}}_l^{UCB}$; $H_l=\emptyset$;
   \FOR{$i\in \mathcal{N}$}
   \IF{$i\notin \tilde{{\bf S}}^l$}
   \STATE $H_l=H_l\cup \{i\}$;
   \ENDIF
   \ENDFOR
   \STATE offer $(\tilde{S}^l_1, \tilde{S}^l_2\cup H_l)$, observe the purchasing decision $c_t=c_t^1\cup c_t^2$; $l_0=l$;
   \REPEAT
   \IF{$c_t^1= \emptyset$}
   \STATE compute $\hat{v}_{i,l}=\sum_{t\in \varepsilon_{l}^1}1(c_t^1=i)$;
   \STATE update $\mathcal{T}^1_i(l)=\{\tau\leq l|i\in S^\tau_1\}$, $T_i^1(l)=|\mathcal{T}^1_i(l)|$, no. of epochs until $l$ that offered product $i$ in the first tier;
   \STATE  update $\bar{v}_{i,l}$ and $v_{i,l}^{UCB}$ according to Eq~\eqref{e.vbar} and Eq~\eqref{e.vucb}; $l=l+1$;
   \STATE compute $\tilde{S}_1^l$ given $\tilde{S}_2^{l_0}$ and ${\bf v}_{l}^{UCB}$ and offer $(\tilde{S}_1^l, \tilde{S}_2^{l_0}\cup H_{l_0})$, observe the purchasing decision $c_t$; $\varepsilon_{l}^1=\varepsilon_{l}^1\cup t$;
   \ELSE 
   \STATE offer $(\tilde{S}_1^l, \tilde{S}_2^{l_0}\cup H_{l_0})$, observe the purchasing decision $c_t$; $\varepsilon_l^1=\varepsilon_l^1\cup t$; $\varepsilon_{l_0}^2=\varepsilon_{l_0}^2\cup t$;
   \ENDIF
   \STATE $t=t+1$;
   \UNTIL{$t=T$ or $c_t=\emptyset$}
   \STATE compute $\hat{v}_{i,l}^{(1)}=\sum_{t\in \varepsilon_l^1}1(c_t^1=i)$;
    $\varepsilon_l^1=\varepsilon_l^1\cup t;$
    \STATE update $\mathcal{T}^1_i(l)=\{\tau\leq l|i\in S^\tau_1\}$, $T_i^1(l)=|\mathcal{T}^1_i(l)|$, no. of epochs until $l$ that offered product $i$ in the first tier; $l=l+1$;
   \STATE   compute $\hat{v}_{i,l}^{(2)}=\sum_{t\in \varepsilon_l^2}1(c_t^2=i)$; $\varepsilon_{l_0}^2=\varepsilon_{l_0}^2\cup t;$
   \STATE update $\mathcal{T}^2_i(l)=\{\tau\leq l|i\in \tilde{S}^\tau_2\cup H_\tau\}$, $T_i^2(l)=|\mathcal{T}^2_i(l)|$, no. of epochs until $l$ that offered product $i$ in the second tier;
   \STATE update $\bar{v}_{i,l}$ and $v_{i,l}^{UCB}$ according to Eq~\eqref{e.vbar} and Eq~\eqref{e.vucb}; $l=l+1$; $t=t+1$;
   \UNTIL{$t=T$}
\end{algorithmic}
\end{algorithm}
We are now ready to present an upper bound on the regret for Algorithm~\ref{A.newproduct}. We provide a proof sketch here and the detailed proof can be found in the Supplementary Material.

\begin{theorem}[Performance bound for Algorithm~\ref{A.newproduct}]
\label{T.regret}
The regret during time $[0,T]$ is bounded above by
\begin{align*}
Reg_\pi(T;\bold{v})&\leq CK\log^2(KT)+C\sqrt{TK\log(KT)}\\
&\hspace{5mm}+ M\sum_{i\in X}v_i (r_{max}-r_i),
\end{align*}
for some constant $C$, where $r_{max}$ is the highest profit of products among $X$, and $K$ is the total number of products.
\end{theorem}


\emph{Proof sketch:} We first rewrite the regret in terms of the epochs. Note that one learning epoch on the secondary tier may correspond to multiple learning epochs on the priority tier. Let $\kappa(l)$ denote as a set of epochs on tier 1 which corresponds to epoch $l\in \mathcal{L}_2$. In Example 2 as shown in Figure 1, we have $\kappa(0)=\{0,1\}$, $\kappa(3)=\{3,4\}$. Thus, the  regret until time $T$ can be expressed
$Reg_\pi(T;\bold{v})=E_\pi[\sum_{l\in \mathcal{L}_2}\sum_{j\in\kappa(l)}\sum_{t\in \varepsilon_j^1}(R_t(\bold{S}_j^*,\bold{v})-R_t((\tilde{S}^j_1, \tilde{S}^l_2\cup H_l),\bold{v}))]$, 
where the set $H_l$ denotes the set of new products with low profit which are added to the second tier at epoch $l\in \mathcal{L}_2$ to satisfy the minimum learning criterion.

Define the ``large probability" event $A_l=\bigcap_{i=1}^K \{  v_{i,l}^{UCB}-C_1\sqrt{\frac{v_i\log(K(l-l_{i,0})+1)}{T_i(l)}}-C_2\frac{\log(K(l-l_{i,0})+1)}{T_i(l)}<v_i<v_{i,l}^{UCB}\}.$ 
Meanwhile, by Lemma~\ref{L.upperbound}, we have $E[R_t(\tilde{\bold{S}}^l,\bold{v})]\leq E[R_t(\bold{S}_l^*,\bold{v})]\leq E[R_t(\bold{S}_l^*,\bold{v}^{UCB})]\leq E[R_t(\tilde{\bold{S}}^l,\bold{v}^{UCB})]$. Thus, conditional on the event $A_l$ and Lemma~\ref{L.upperbound}, we can show that  $E[R_t(\bold{S}_l^*,\bold{v})-R_t((\tilde{S}_1^l, \tilde{S}_2^l\cup H_l),\bold{v})]$ can be bounded above by $E[R_t(\tilde{\bold{S}}^l,\bold{v}_l^{UCB})-R_t(\tilde{\bold{S}}^l,\bold{v})]+E[R_t(\tilde{\bold{S}}^l,\bold{v})-R_t((\tilde{S}_1^l,\tilde{S}_2^l\cup H_l),\bold{v})]$.  

We see that the regret consists of two parts: The first term can be bounded above by \\
$E[\sum_{l\in \mathcal{L}_1}\sum_{i\in \tilde{S}_1^l}r_i(v_{i,l}^{UCB}-v_i)+\sum_{l\in \mathcal{L}_2}\sum_{i\in\tilde{S}_2^l}r_i(v_{i,l}^{UCB}-v_i)].$ The second term can also be bounded since each product will be included in the set $H$ for at most $M$ times. 

Combined this result on the ``large probability'' event $A_l$ with the error on the measure of ``small probability" event $A_l^c$, the upper bound of the regret can be obtained. $\blacksquare$


We want to point out Algorithm~\ref{A.newproduct} can be easily extended to include more than two tiers, and Theorem \ref{T.regret} will continue to hold. The regret bound in Theorem \ref{T.regret} consists of three terms, where the first two terms account for the estimation error on product valuation, while the third term is linear with $M$, representing the price one has to pay in order to include new products with low profit into learning. When $M=0$, Theorem \ref{T.regret} provides the regret bound for the case \emph{without} the minimum learning criterion, which is a special case discussed in Section \ref{sect:nocheapproduct}.

\section{Numerical experiments}
In this section, we conduct three experiments. We first investigate the robustness of Algorithm~\ref{A.newproduct}. Next, we compare Algorithm~\ref{A.newproduct} that simultaneously explores and exploits with a benchmark algorithm which separates the two phases. Lastly, we compare our algorithm with an alternative strategy for learning new products. 

\paragraph{Experiment 1 (Robustness study)} \label{ss.robust} We consider a setting where $X$ contains 80 products with profit $r_i$ uniformly distributed on [0,1] and 20 products with $r_i$ uniformly distributed [0,0.2]. We compare four scenarios, when the product valuation $v_i$ is uniformly distributed on [0,0,1], [0,0.2], [0,0.3], and [0,0.5]. A new product is introduced after every 800 time steps.  We set $M=100$ for the minimum learning criteria. 

\begin{figure}
\centering
  \includegraphics[width=0.6\linewidth]{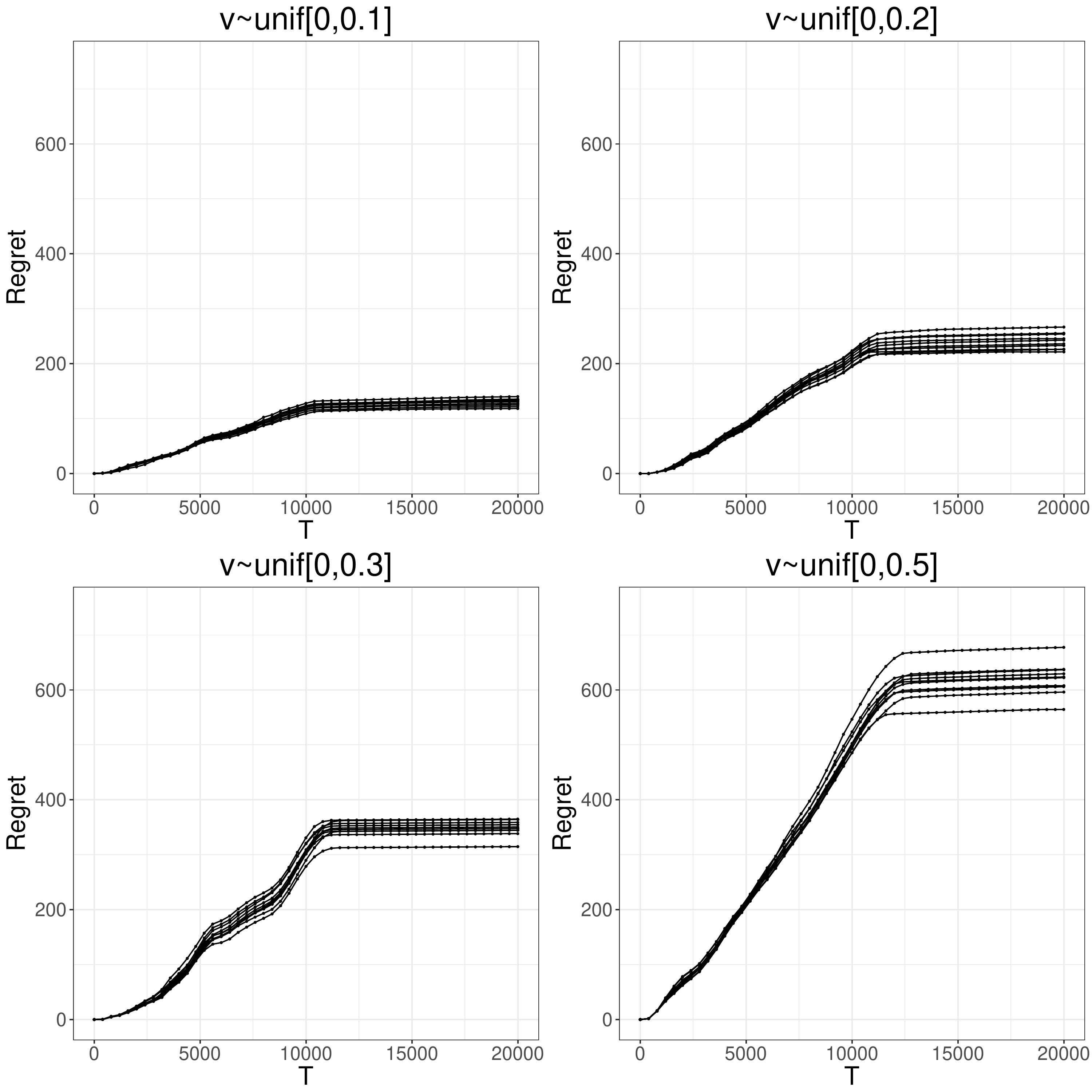}
  \caption{Comparison of regrets generated under Algorithm~\ref{A.newproduct} for four different scenarios. 
  }
  \label{fig:regret1}
\end{figure}

Figure~\ref{fig:regret1} shows the results based on 10  independent simulations for different distributions of $\bold{v}$. The average regrets are 129.87, 243.38, 348.31, and 620.14 for the four scenarios. Notice that both the mean and variance of the regret are increasing with the support of $\bold{v}$. It implies that the learning process is harder when the product valuations $\bold{v}$ lie on a larger support and have  higher variability.

\paragraph{Experiment 2 (Comparison with a explore-then-exploit benchmark)} The benchmark we consider is adapted from \citealp{saure2013optimal}. As shown in Section~\ref{ss.optimal}, there are at most $|X_1||X_2|$ candidates which are profit-ordered by tier. In the exploration phase of the benchmark algorithm,  every candidate whose profit is higher than the current optimum is offered for at least $\gamma\log(t)$ times, where $\gamma$ is a tuning parameter. In the exploitation phase, the algorithm uses the estimated parameters to determine a tiered offering with the highest expected profit and offer it to all customers.

For the experiment, consider the setting that $X$ contains 12 products, where the profit $r_i$ of 8 of them are  uniformly distributed on [0,1], and that of 4 products on [0,0.2]. The valuation $v_i$ is uniformly distributed on [0,0.1]. For ease of  comparison, all products are launched at $t=0$. Set $M=100$.

\begin{figure}
\centering
  \includegraphics[width=0.7\linewidth]{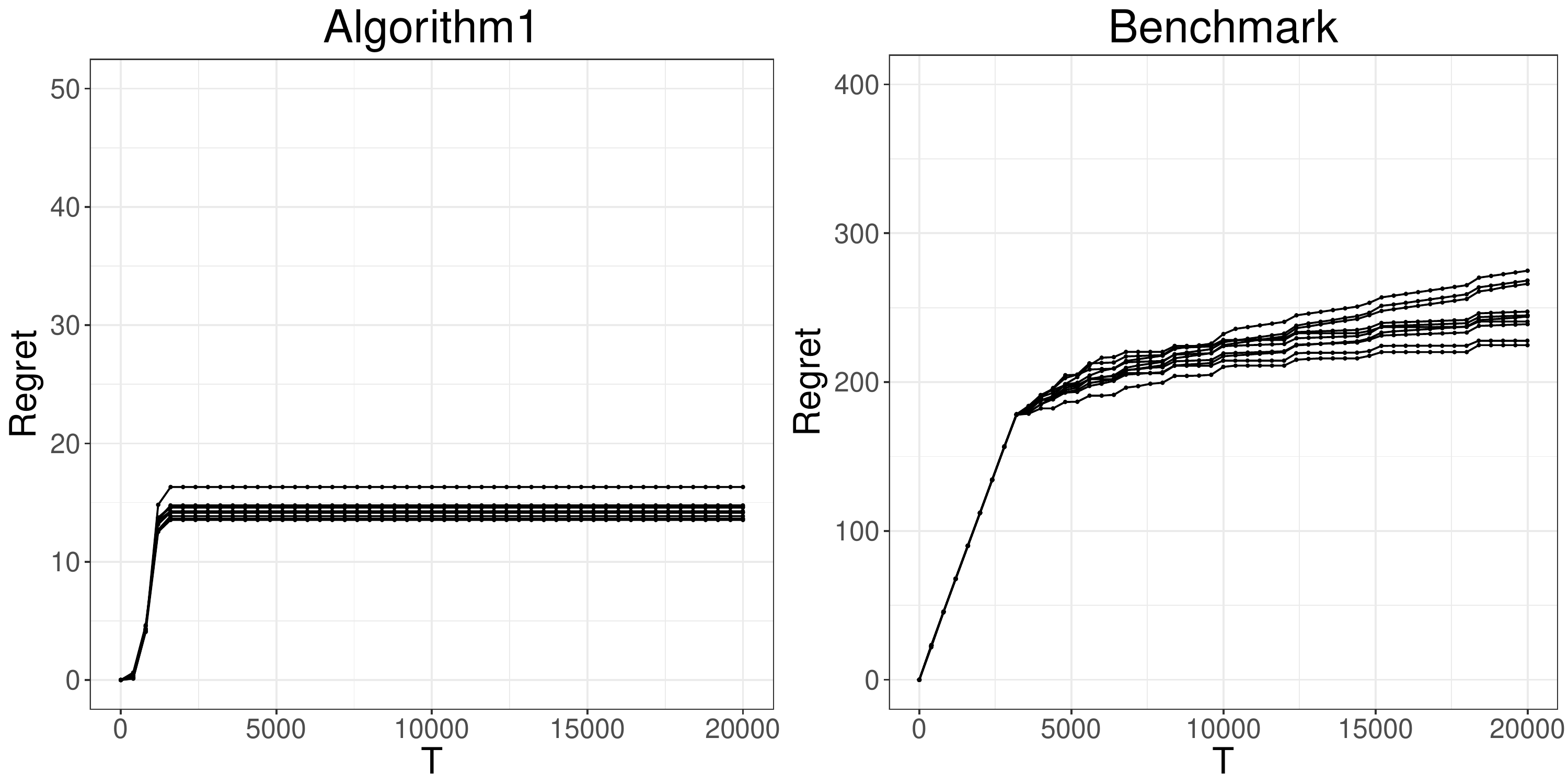}
  \caption{Comparison of Algorithm~\ref{A.newproduct} with an explore-then-exploit benchmark algorithm.}
  \label{fig:regret2}
\end{figure}

Figure~\ref{fig:regret2} shows the results based on 10 independent simulation. It depicts the superiority of our algorithm  over the benchmark, where the average regrets are 14.39 and  247.78 under  Algorithm~\ref{A.newproduct} and the benchmark respectively. 

\paragraph{Experiment 3 (Comparison with an alternative learning strategy for new products)} We have shown in Algorithm~\ref{A.newproduct} that new products with profit lower than $E[R(S_2^*)]$ will be added to the secondary tier. In this experiment, we  compare it with an alternative strategy where those new products with low profit will be randomly added to either tier with equal probability for learning. To be precise, we consider a setting where $X_1$ contains 20 products with profit uniformly distributed on [0.5,1] and valuation on [0,0.1]. $X_2$ contains 30 products with profit uniformly distributed on [0,0.6] and valuation on [0,0.2]. We compute the optimal product offering as the current offering based on these values. Next, we assume 15 new products with profit uniformly distributed on [0,0.55] and valuation on [0,0.3] are launched at time $t=0$. For the benchmark, new products with profit below $E[R(S_2^*)]$ will be randomly added to one of the tiers. Set $M=300$. 

\begin{figure}
\centering
  \includegraphics[width=0.7\linewidth]{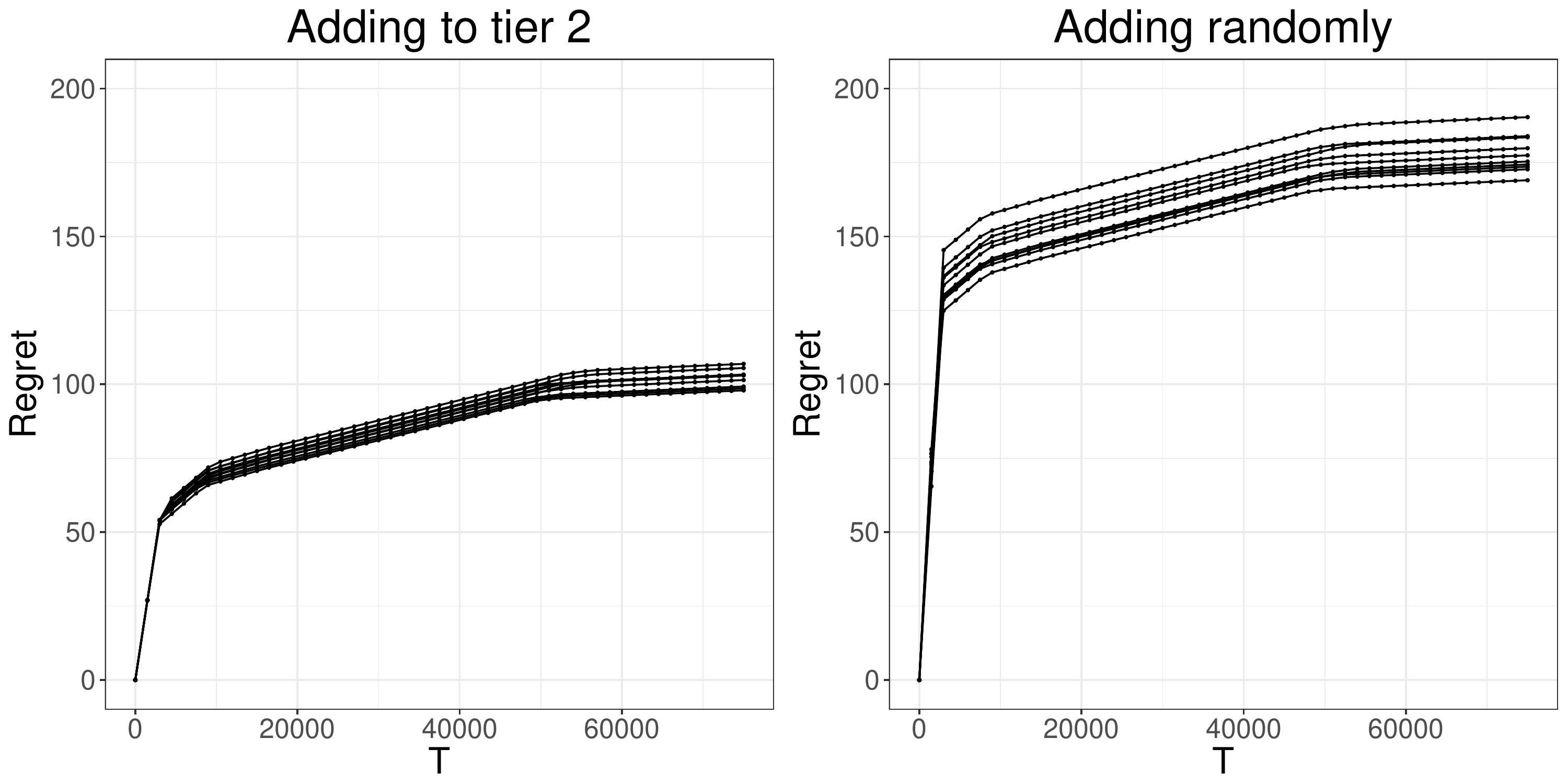}
  \caption{Comparison with an alternative learning strategy for new products. }
  \label{fig:regret3}
\end{figure}

As shown in Figure~\ref{fig:regret3}, the average regrets are 102.21 under Algorithm~\ref{A.newproduct} and 178.00 under the alternative strategy.  It highlights the benefit of having a tiered offering as one could use the secondary tier to mitigate some profit risk when learning with new products.


\section{Conclusion}
In this work, we studied a product selection problem with a SMNL model which specifies the order in which products are being presented.  For the offline setting where the product valuations are known, a polynomial-time solvable algorithm was provided. For the online setting, we analyzed a novel setup where multiple new products could arrive in  the middle of a selling period. 
Depending on the presence of the minimum learning criterion, we proposed an online  algorithm and characterized its regret.

There are several future directions of this work. For instance, products' valuations may vary with time, especially for fashion and technology products. Thus, there is a need for an online algorithm that learns the dynamic valuations. In addition, it {would} be interesting to utilize customer attribute data and historical sales data to provide personalized recommendations. 

\nocite{langley00}

\bibliography{fatigue}
\bibliographystyle{apalike}

\end{document}